%% file: arxiv.tex
\documentclass[10pt,twocolumn,letterpaper]{article}

\usepackage[pagenumbers]{iccv} 
\usepackage{multirow}

\definecolor{iccvblue}{rgb}{0.21,0.49,0.74}
\usepackage[pagebackref,breaklinks,colorlinks,allcolors=iccvblue]{hyperref}

\def\paperID{7457} 
\def\confName{ICCV}
\def\confYear{2025}

\title{Stepping Out of Similar Semantic Space for Open-Vocabulary Segmentation}

\author{Yong Liu\textsuperscript{1}\footnotemark[1]~,
        Songli Wu\textsuperscript{1}\footnotemark[1]~,
        Sule Bai\textsuperscript{1}\footnotemark[1]~,
        Jiahao Wang\textsuperscript{2}~,
        Yitong Wang\textsuperscript{3}~,
        Yansong Tang\textsuperscript{1}\footnotemark[2]~\\
\textsuperscript{1}Tsinghua Shenzhen International Graduate School, Tsinghua University\\
\textsuperscript{2}The University of Hong Kong~
\textsuperscript{3}ByteDance Inc.\\
{\tt\small liuyong23@mails.tsinghua.edu.cn, tang.yansong@sz.tsinghua.edu.cn}
}

\begin{document}
\maketitle
\footnotetext[1]{Equal contribution}. 
\footnotetext[2]{Corresponding author}

\input{sec/0_abstract}    
\input{sec/1_main}
{
    \small
    \bibliographystyle{ieeenat_fullname}
    \bibliography{main}
}

\end{document}

%% file: sec/0_abstract.tex
\begin{abstract}
Open-vocabulary segmentation aims to achieve segmentation of arbitrary categories given unlimited text inputs as guidance. To achieve this, recent works have focused on developing various technical routes to exploit the potential of large-scale pre-trained vision-language models and have made significant progress on existing benchmarks. However, we find that existing test sets are limited in measuring the models' comprehension of ``open-vocabulary" concepts, as their semantic space closely resembles the training space, even with many overlapping categories.
To this end, we present a new benchmark named OpenBench that differs significantly from the training semantics.
It is designed to better assess the model's ability to understand and segment a wide range of real-world concepts. When testing existing methods on OpenBench, we find that their performance diverges from the conclusions drawn on existing test sets.
In addition, we propose a method named OVSNet to improve the segmentation performance for diverse and open scenarios. Through elaborate fusion of heterogeneous features and cost-free expansion of the training space, OVSNet achieves state-of-the-art results on both existing datasets and our proposed OpenBench.
Corresponding analysis demonstrates the rationality and effectiveness of our proposed benchmark and method.
\end{abstract}

%% file: sec/1_main.tex
\vspace{-10pt}
\section{Introduction}
\label{sec:intro}
Semantic segmentation is one of the most fundamental tasks in computer vision, which targets at assigning semantic category to pixels in an image. Despite achieving excellent progress in recent years~\cite{sca,qdmn,gsfm,segformer, deeplab, peftris, qpmn,unilseg, qdmn++,yuji1,yuji2,univg,hyperseg,hdc,instructseg}, traditional semantic segmentation approaches rely on predefined categories sets and tend to falter when encountering categories absent during the training phase, significantly impeding their real-world application. 
Such challenge has inspired the exploration of Open-Vocabulary Segmentation (OVS) task~\cite{zegformer,gkc,pmosr,PAD,openseg,spnet,lseg, scan}. 
Different from traditional closed-set segmentation, OVS methods aim to segment arbitrary categories under the guidance of given text inputs, 
which has many potential applications such as auto-driving and human-robot interaction~\cite{autodrive,thinkbot}.

\begin{figure}[t]
    \centering
    \includegraphics[width=\linewidth]{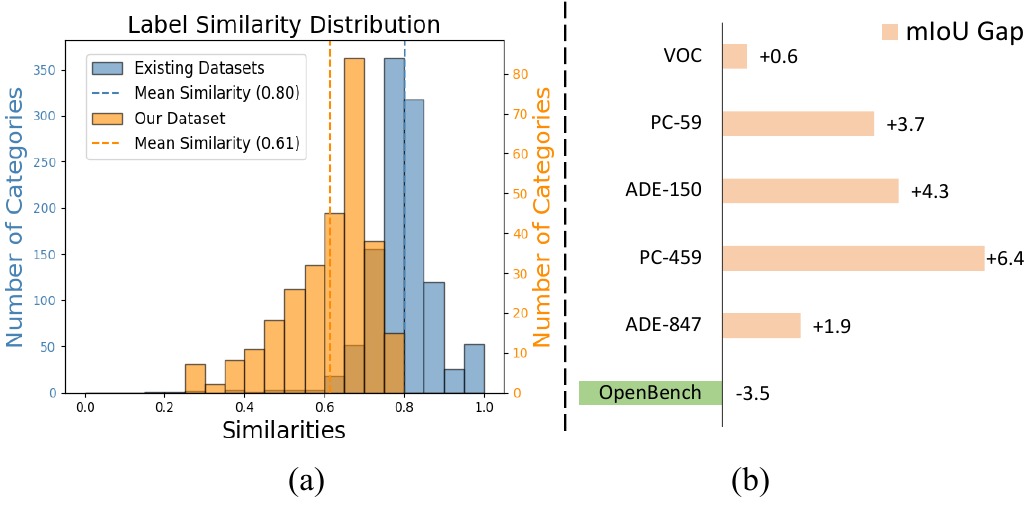}
    \vspace{-25pt}
    \caption{(a) shows the similarity distribution of the existing test set (blue) and our OpenBench (orange) with the training semantic space. (b) illustrates the performance gap between the methods finetuning~\cite{catseg} and frozen~\cite{san} CLIP on different benchmarks.}
    \label{fig:teaser}
    \vspace{-15pt}
\end{figure}

However, it is challenging to identify unseen categories without the intervention of external knowledge. 
Therefore, an intuitive idea is to introduce large-scale vision-language model~\cite{clip, align} trained with numerous sources to extend the semantic space of segmentation models. 
Following this idea, existing OVS methods~\cite{catseg,sed,scan,san,masqclip,adapt-mask} concentrate on exploring how to exploit the pretrained knowledge prior more effectively and 
derive several technical routes.
Early methods~\cite{Simbaseline,zegformer,adapt-mask} adopt a two-stage pipeline. This approach first generates class-agnostic mask proposals with segmentation models, following which the pre-trained CLIP~\cite{clip} serves as an additional classifier to execute region-level open-vocabulary classification. However, this strategy incurs heavy computation overhead, as it requires repeated forward passes of the CLIP vision encoder for each mask. This can be prohibitively expensive for real-world applications. 
To address this problem, researchers~\cite{san,maft,masqclip,fcclip} propose to unify classification and segmentation within a shared space, \textit{i.e.}, integrating frozen CLIP features into the segmenter and extracting CLIP embeddings from potential target regions to enable simultaneous classification and segmentation. 
These methods achieve remarkable  performance in a more efficient one-stage pipeline. 
Subsequently, some works~\cite{catseg,sed} discover that finetuning the CLIP encoder could further enhance the model's performance on existing test sets. They develop an early fusion paradigm based on cost map and achieve outstanding segmentation results.
However, this performance boost is somewhat counterintuitive, as fine-tuning CLIP typically adapts it to specific training semantic spaces, which could reduce its original generalization ability. Yet, the ability of generalized visual-linguistic alignment is precisely what the open-vocabulary task requires. So why does fine-tuning CLIP appear to benefit the performance of OVS?



This question drives us to revisit the existing evaluation benchmarks of the cross-dataset evaluation paradigm. We count the maximum similarity of each category in test set to the training semantics. 
\Cref{fig:teaser} (a) illustrates the similarity distribution of all test categories to the training space (blue color). 
Results indicate that the existing test data is, in fact, highly similar to the training semantic space.
Specifically, even the most challenging ADE-847~\cite{ade20k} and PC-459~\cite{pascal} datasets have an average similarity to the training classes of 0.79 and 0.83. PC-59~\cite{pascal} and VOC~\cite{pascal-voc} datasets even have a similarity to the training semantics of 0.95 and 0.97. 
This explains why finetuning CLIP on training set can lead to further performance gains on existing benchmarks. 
However, it somewhat conflicts with the purpose of the open-vocabulary setting, which aims to achieve segmentation of unlimited categories in the real world.

To address the above problem and provide a more comprehensive measure of OVS models, we propose a new evaluation benchmark named  OpenBench, which is characterized by significant semantic differences from the training data. \Cref{fig:teaser} (a) also demonstrates the similarity distribution of our OpenBench (orange color). It can be seen that our benchmark contains more novel classes compared to the existing test sets. 
We evaluate the performance of approaches across different technical routes on our benchmark, and the findings diverge from those obtained on existing test sets.
As shown in \Cref{fig:teaser} (b), while finetuning CLIP yields steady performance gains on existing datasets, it shows a significant drop on our benchmark. This outcome also supports our motivation for proposing a benchmark that diverges substantially from the training space.

In addition, we also present a framework named OVSNet to improve the segmentation performance for diverse and open scenarios. 
In OVSNet, we adopt a Proxy Calibration (PC) training strategy to enhance the robustness of model representations by broadening the training space with no additional cost. This strategy leverages the synthesis of proxy embeddings, which approximate unseen semantics through convex combinations of in-vocabulary classes.
Additionally, inspired by random walk~\cite{randwalk}, we propose a gradient-free fusion of CLIP embeddings and segmentation decoder features to model a robust joint space and mitigate potential overfitting issues related to training semantics.
With the above proposed designs calibrating model knowledge from the aspects of training space and features separately, our OVSNet demonstrates strong performance across a wide range of scenarios, both on existing test sets similar to the training semantics and on our proposed novel OpenBench. 
Related ablations also demonstrate the effectiveness and soundness of our proposed strategies.

Our contributions can be summarized as follows:
\begin{itemize}
    \item We present a new benchmark named  OpenBench for OVS, which is characterized by significant semantic differences from the training space. 
    While using it to evaluate the comprehension of OVS methods in more open scenarios, we reach a different conclusion from that of the existing test sets.
    \item We propose OVSNet to improve the segmentation performance on diverse and open scenarios. It adopts a Proxy Calibration training strategy to broaden training space without additional cost. Besides, a gradient-free algorithm is leveraged to fuse CLIP embeddings and segmentation features for modeling robust joint space.
    \item Our method achieves state-of-the-art performance on existing benchmarks and the proposed OpenBench. Extensive experiments are conducted to prove the effectiveness and soundness of the proposed algorithms.
\end{itemize}

\begin{figure*}[t]
    \centering
    \includegraphics[width=\linewidth]{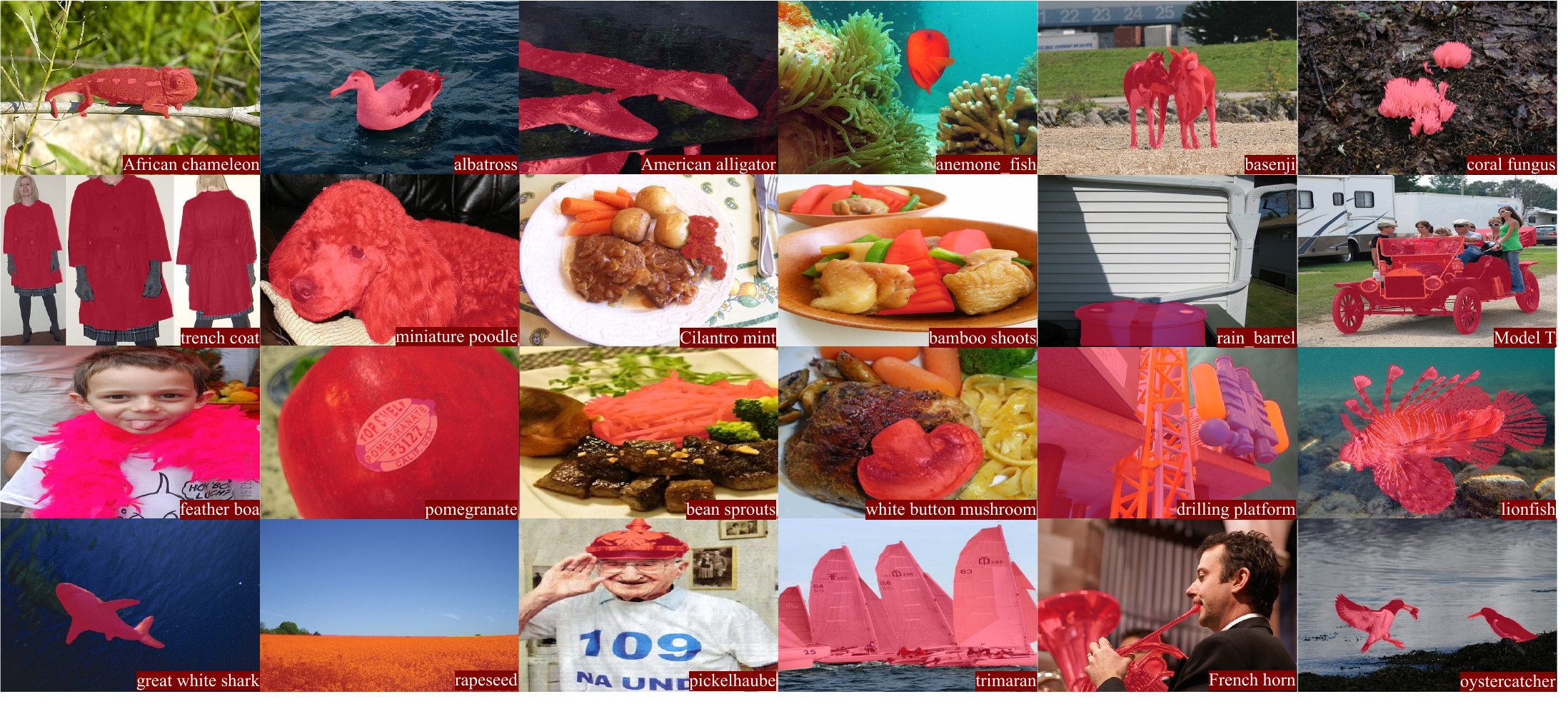}
    \vspace{-20pt}
    \caption{Illustration of the samples in the proposed OpenBench. Compared to existing test sets, our OpenBench has semantic categories that differ more from the training space, which helps to provide a more comprehensive measure of the model's ability to perform open-vocabulary segmentation. The corresponding category name is labeled in the lower right corner of each image.}
    \label{fig:dataset_show}
\end{figure*}

\input{tables/dataset}

\section{Related Work}
\paragraph{Open-Vocabulary Segmentation.}
This task aims to segment an image and identify regions with arbitrary given text queries~\cite{openseg,zs3net, soc, spnet, scclip}.
Pioneering work~\cite{spnet} replaces the output convolution layer by computing the similarity between visual features and linguistic embeddings, which has become common practice.
Later, some methods~\cite{openseg,Simbaseline,zegformer,adapt-mask, odise} propose a two-stage pipeline: the model first generates class-agnostic mask proposals, then 
 a pretrained CLIP~\cite{clip} is utilized to perform sub-image classification by cropping and masking corresponding regions. With the combination of both in-vocabulary and out-vocabulary classification, these methods obtains great performance.
 However, this strategy incurs heavy computation overhead, as it requires repeated forward passes of the CLIP vision encoder for each mask. This can be prohibitively expensive for real-world applications. 
To address this problem, subsequent researchers propose to unify classification and segmentation within a shared space. 
Among them,
SAN~\cite{san} designs a side-adapter network to leverage CLIP features for reorganizing segmentation and classification.
GKC~\cite{gkc} presents text-guided knowledge distillation strategy to transfer CLIP knowledge to specific classification layer.
FCCLIP~\cite{fcclip} proposes to leverage the local-awareness advantage of convolution backbones.
Some works~\cite{catseg,sed} then discover that finetuning CLIP encoder could further enhance the model's performance on existing test sets and develop effective one-stage framework.

However, we find that existing test sets are highly similar to the training set, which limits their ability to effectively evaluate the model’s comprehension of the open vocabulary semantics. Therefore, in this paper we propose a new benchmark OpenBench that is characterized by significant semantic differences from the training data.  We also propose a method named  OVSNet  to improve the segmentation performance for diverse and open scenarios.

\vspace{-10pt}
\paragraph{Vision-Language Pre-training.}
Vision-language pre-training aims to learn a unified visual-linguistic representation space. Early approaches~\cite{pretrain1, pretrain2, pretrain3, pretrain4, qu2022siri,haoji1,haoji2,flashvstream}, constrained by small-scale datasets, struggle to achieve strong performance and require fine-tuning for downstream tasks. However, with the advent of large-scale web data, recent works~\cite{pretrain5, clip} have demonstrated the benefits of leveraging such data to train a more robust multi-modal representation space. Among these, CLIP~\cite{clip}, the most widely used vision-language model, employs contrastive learning to link images with their corresponding captions, achieving impressive cross-modal alignment. Building upon previous works~\cite{zegformer, lseg, Simbaseline, oclip, proxydet}, we also utilize the well-aligned and generalized space of CLIP to enhance open-vocabulary segmentation.

\section{Benchmark}
\subsection{Dataset Introduction}
In \Cref{tab:dataset_show}, we present the statistics comparison of our proposed OpenBench with existing popular open-vocabulary segmentation datasets, including ADE-150~\cite{ade20k}, ADE-847~\cite{ade20k}, PC-59~\cite{pascal}, PC-459~\cite{pascal}, VOC~\cite{pascal-voc}, and Cityscapes~\cite{cityscape}. Specifically, our OpenBench contains 286 categories, of which the maximum similarity to the training categories is 0.7947, the minimum is 0.2608, and the average similarity is 0.6142. 
It can be observed that, compared to existing datasets, our OpenBench exhibits a greater disparity from the training set, which better evaluates the model's segmentation ability for arbitrary open categories.


Besides, our OpenBench contains fine-grained categories, \textit{e.g.}, ``oystercatcher",  rather than just generic broad category such as ``bird” (corresponds to the ``Fine Granularity” characterization in \Cref{tab:dataset_show}). This is important for evaluating whether the OVS models preserve the generalization capability of CLIP~\cite{clip}, which is inherently capable of detecting these fine-grained categories. 
Actually, PC-459~\cite{pascal} and ADE-847~\cite{ade20k} also have fine-grained categories, but they suffer from semantic duplication problem. 
In detail, semantic duplication means that there will be both fine-grained categories as well as corresponding broad categories in these datasets, which would cause bias in the metrics calculation.
For instance, if a model assigns the ``door'' tag to a region whose ground truth label is ``double door", it will be considered incorrect.
For open-vocabulary segmentation setting, we argue that the responsibility of the model is to discern the correct semantic.
The distribution of such duplication categories and the corresponding performance statistics is unreasonable.
SCAN~\cite{scan} also points similar observations and try to solve this by designing new metric. However, this requires manual judgment of the relationships between categories, introducing additional cost and uncertainty. 
Our OpenBench, on the other hand, has fine-grained semantics without the problem of semantic duplication, which helps to more accurately reflect the real capabilities of related models.

\Cref{fig:dataset_show} illustrates part of the categories contained in OpenBench, where some categories such as food related even differ significantly from the visual domain of the training set.
Furthermore, our OpenBench also accounts for ``other" or ``background" pixels in images, which are often overlooked by existing datasets.
This is important for open-vocabulary segmentation because the model is essentially forced to choose the most appropriate category from a predefined set. If there is no correct class and no ``others" option in the candidate set, the model would be compelled to make an incorrect prediction. 
Moreover, ``others" or ``background" require the model to consider not only the direct visual-linguistic correspondence between the image content and each class, but also the relative relationships among different classes. This adds complexity and makes the task more relevant to real-world scenarios.

\subsection{Dataset Collection}
Our OpenBench is primarily derived from existing accessible segmentation datasets. 
To obtain semantic categories that differ significantly from the training set, \textit{i.e.}, COCO~\cite{coco}, we first organize a group of segmentation datasets that possess category annotations, \textit{e.g.}, Food103~\cite{food103}, Imagenet-S~\cite{imagenet-s}, and CamVid~\cite{camvid}. 
For all categories in the collected data, we leverage CLIP-L/14~\cite{clip} to compute their maximum similarity with COCO categories. 
Then we calculate the minimum similarity among the categories present in each image as the image's similarity to the training semantics.
If the image similarity is greater than the set threshold $\sigma_1$, the image will be filtered. 
Conversely, if the similarity of the image is less than the threshold $\sigma_1$, we would set the category exist in the image whose similarity is greater than the threshold $\sigma_2$ to ``others” and save the related images and annotation. Finally, after manually filtering the categories for potential conflicts or duplicates, we get the OpenBench.

\begin{figure*}[t]
    \centering
    \includegraphics[width=\textwidth]{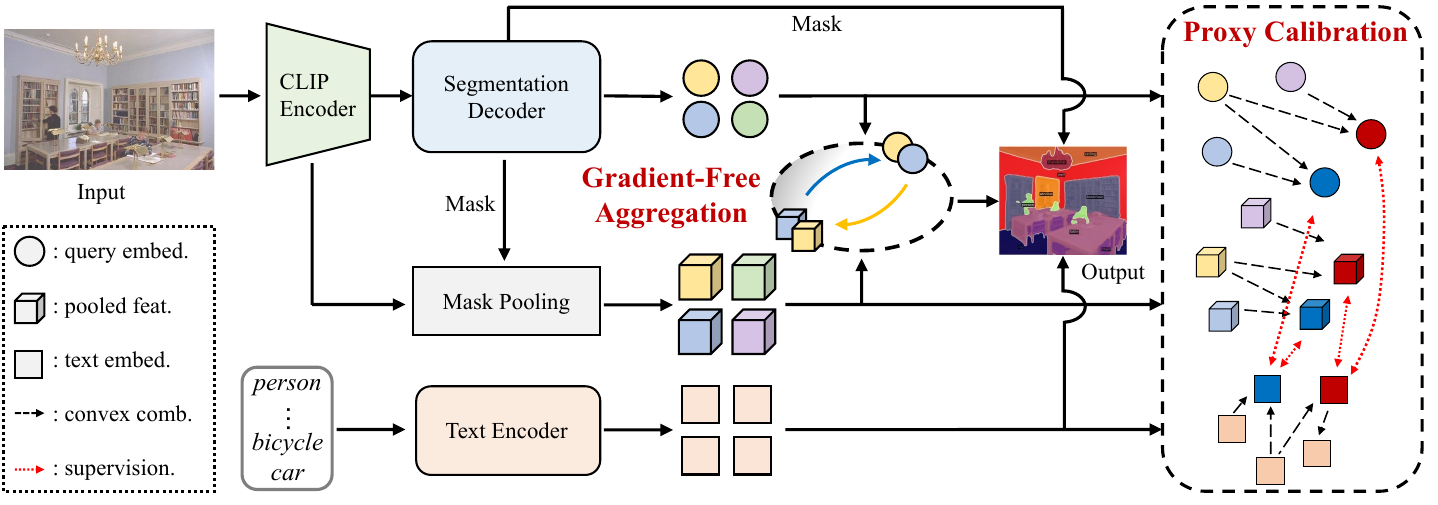}
    \vspace{-15pt}
    \caption{Pipeline of our OVSNet. The input image and corresponding categories are first encoded by pretrained CLIP~\cite{clip}. A segmentation decoder is trained to decode the visual features and generate potential masks as well as corresponding query embeddings. We then leverage the predicted masks to extract related CLIP features with Mask Pooling operation. To integrate different knowledge spaces and improve the representation robustness, Gradient-Free Aggregation fuses the query embeddings with CLIP features. The aggregated visual embeddings are combined with class embeddings to perform visual-linguistic alignment and generate final output. During training, we apply Proxy Calibration strategy to expand training space in a cost-free manner.}
    \label{fig:pipeline}
    \vspace{-5pt}
\end{figure*}

\section{Method}
\subsection{Overview}
The overall pipeline of our proposed OVSNet is shown in \Cref{fig:pipeline}. 
We leverage the vision and text encoder of pretraind CLIP~\cite{clip} to encode the input image and category names, respectively. 
Then, a segmentation decoder is trained to decode the visual features, generating masks for potential targets and their corresponding query embeddings.
We then leverage the predicted masks to extract related CLIP features with Mask Pooling operation. To integrate heterogeneous knowledge spaces and improve the representation robustness, Gradient-Free Aggregation fuses the query embeddings with pooled CLIP features. The aggregated visual embeddings are combined with linguistic embeddings to perform visual-linguistic alignment and generate final output. During training, a Proxy Calibration strategy is utilized to expand training space by applying supervision on the convex combination of training semantics.

\subsection{Gradient-Free Aggregation}
Although CLIP~\cite{clip} exhibits strong generalization ability, it optimizes for image-level visual-linguistic alignment during pretraining, which creates a great domain gap compared to mask-level alignment. As a result, directly using CLIP features obtained through mask pooling for mask classification may lead to suboptimal performance.
The query embeddings generated by the segmentation decoder, on the other hand, have a strong region-level alignment prior owing to their suitability for the training space. However, they struggle to recognize novel semantics.

Therefore, we propose to reorganize them to obtain more robust domain-adapted representations.
An intuitive approach is to learn the intrinsic connection between them through self-attention or cross-attention mechanisms. 
However, we observe that such learning-based strategy tends to develop excessive reliance on query embeddings fitted to seen semantics during the training process while neglecting the utilization of more generalizable information provided by CLIP, ultimately affecting the model's ability to understand unlimited categories.
Therefore, we propose to combine CLIP features and query embeddings via a gradient-free approach to avoid potential overfitting problems during the learning of the joint space. This is inspired by Random Walk algorithm~\cite{randwalk, uvcom}. 
Specifically, there are multiple iterations in this process where two features learn collaboratively in shared embedding space until convergence.
 Suppose we have $N$ queries. For convenience, we omit the query index and take a random query as the example.
The corresponding query embeddings and CLIP features can be represented by $F_Q$ and $F_C$.
$F_Q$ and $F_C$ are defined as the initial state $F_Q^0$ and $F_C^0$ at the $0$-th iteration. Then we formulate affinity $\mathcal{Z}$ by scaled dot product: $\mathcal{Z}=\lambda F_C^0 (F_Q^0)^{\top }$, where $\lambda$ is the scaling factor.
At $t$-th iteration, the query embedding $F_Q^t$ is derived by the original embedding $F_Q^0$ and the updated CLIP feature $F_C^{t-1}$ from previous iteration, which can be formulated as:
\begin{equation}
    \label{txt_t_iter}
    F_Q^{t}=\omega \mathrm{Norm}(\mathcal{Z})^{\top }F_C^{t-1}+(1-\omega )F_Q^{0},
\end{equation}
Subsequently, it is utilized to boost the CLIP feature of this iteration:
\begin{equation}
    \label{vid_t_iter}
    F_C^{t}=\omega \mathcal{Z}F_Q^{t} + (1-\omega )F_C^{0},
\end{equation}
where $\omega \in (0,1)$ is the factor that controls the degree of feature fusion.
By combining the above equations we can obtain the overall iterative update formula of $F_C^{t}$:
\begin{equation}
    F_C^{t} = (\omega^2A)^{t} F_C^{0} + (1-\omega )\sum_{i=0}^{t-1} (\omega^2A)^{i}(\omega \mathcal{Z}F_Q^{0}+F_C^{0}),
\end{equation}
where $A$ denotes $\mathcal{Z}\mathrm{Norm}(\mathcal{Z})^{\top}$. 
Moreover, to avoid the potential issue of unexpected gradient and simplify the calculation process, we use an approximate inference function based on Neumann Series~\cite{neuman} when $t \to \infty$. Then the update formula can be present as:
\begin{equation}
    \label{new_vid}
    F_C^{\infty}=(1-\omega )(I-\omega ^2 A)^{-1}(\omega \mathcal{Z} F_Q^{0}+ F_C^{0}),
\end{equation}
where $I$ represents the identity matrix.
With this gradient-free aggregation strategy, the model can recombine query embeddings with CLIP features to produce more robust region-level representations.

\subsection{Proxy Calibration}
A broader semantic training space tends to result in more generalized model representations.
However, expanding the training space for OVS is costly due to the intensive prediction nature of segmentation tasks. To alleviate this problem, we propose a cost-free approach named Proxy Calibration (PC) to expand the training space and  enhance the generalization ability of the model. It leverages the synthesis of proxy embeddings, which approximates novel semantics via convex combination between the in-vocabulary classes.

\input{tables/main_results}

Specifically, having the aggregated query embeddings $F_{Q}$, we perform random convex combination of them during training to simulate embedding space for undefined classes, as shown by the black dotted line in \Cref{fig:pipeline}. 
This process can be formulated as:
\begin{equation}
    F'_{Qmn} = \alpha * F_{Qm} + (1 - \alpha) * F_{Qn}
\end{equation}
where $m, n \in \{1, 2, ..., N\}$. $\alpha$ is a random weight sampled from the distribution of $Beta(\gamma, \gamma)$. $F'$ denotes the generated proxy embeddings.
The same enhancement is performed for the corresponding CLIP feature $F_{C}$ and linguistic embeddings $F_{T}$.
Note that the text features here refer to the class representations of the ground truth regions corresponding to each query, which are obtained after Hungarian Matching~\cite{detr}.
Having the generated $F'_Q$, $F'_C$, and $F'_T$, we perform distance supervision on them to promote region-level visual-linguistic alignment performance from the perspective of proxy semantics. That is:
\begin{align}
    &\mathcal{L}_{PQ} = 1 - \frac{\mathbf{F'_Q} \cdot \mathbf{F'_T}}{\|\mathbf{F'_Q}\|_2 \|\mathbf{F'_T}\|_2}, \\
    &\mathcal{L}_{PC} = 1 - \frac{\mathbf{F'_C} \cdot \mathbf{F'_T}}{\|\mathbf{F'_C}\|_2 \|\mathbf{F'_T}\|_2},
\end{align}
where $\mathcal{L}_{PQ}$ and $\mathcal{L}_{PC}$ denote the applied proxy supervision for query embeddings and pooled CLIP features, respectively. The final proxy loss is the sum of $\mathcal{L}_{PQ}$ and $\mathcal{L}_{PC}$.
By applying additional supervision in this way, the model is not limited to a specific training space and can learn more robust visual representations.

\section{Experiment}
\subsection{Evaluation Benchmark and Metric}

To evaluate the effectiveness and generalization ability of our method on diverse scenarios, we conduct extensive experiments on the popular existing benchmarks, ADE20K150~\cite{ade20k}, ADE20K847~\cite{ade20k}, Pascal VOC~\cite{pascal-voc}, Pascal Context-59~\cite{pascal}, Pascal Context-459~\cite{pascal} and our proposed OpenBench.
ADE20K is a large-scale scene understanding benchmark, containing 20k training images, 2k validation images, and 3k testing images.
There are two splits of this dataset. 
ADE20K-150 contains 150 semantic classes whereas ADE20K-847 has 847 classes. The images of both are the same.
Pascal Context is an extension of Pascal VOC 2010, containing 5,005 validation images. 
We take the commonly used PC-59 and challenging PC-459 version for validation.
Pascal VOC contains 11,185 training images and 1,449 validation images from 20 classes. We use the provided augmented annotations.
Following previous works~\cite{Simbaseline,zegformer,openseg}, we take the \textit{mean-intersection-over-union} (mIoU) as the metric to compare our model with previous state-of-the-art methods.
In ablation, we assume the use of base-level CLIP as the backbone for comparison.

\subsection{Implementation Details}
We utilize the Mask2Former~\cite{video_mask2former} as the segmentation decoder and the implementation is based on {\tt detectron2}~\cite{wu2019detectron2}.
The model is trained with batch size of 8 and total iteration of 60k.
The base learning rate is 0.0001 with a polynomial schedule. 
Following previous works~\cite{maftplus,fcclip}, we take CLIP~\cite{clip} of ConvNeXt as the backbone.
The  input image is resized and cropped to 1024$\times$1024. 
For data augmentation, random flip and color augmentation are adopted. The weight decay of the model is 0.05. 
$\lambda$ and $\gamma$ of GFA and PC are set to 0.2 and 2 by default.
The segmentation loss consists of dice loss and cross entropy loss. The classification loss is cross entropy loss.
For the weights of the loss function, we set 5 and  2 for segmentation loss and classification loss, respectively.   Other hyperparameters are the same as Mask2Former~\cite{cheng2021mask2former}.

\subsection{Main Results}
We compare our model with existing state-of-the-art approaches in \Cref{tab:main_result}. 
It can be seen that with proxy calibration expanding training semantics and the gradient-free aggregation algorithm modeling the joint space, our model achieves the state-of-the-art performance on most popular benchmarks.
\textit{The last column counts the average performance across multiple datasets, which shows that our method performs well for both in-vocabulary as well as out-vocabulary diverse scenarios.} 
The relatively inferior performance of our method on VOC is due to the fact that the semantics of this dataset are almost contained in the training set, which tends to benefit models such as SCAN~\cite{scan} that finetunes the CLIP backbone on training set.
Besides, the greater divergence of OpenBench from training semantics does not imply that it is totally more challenging than other benchmarks, because the final performance is influenced by various factors such as image content complexity, image clarity, and the number of given categories. More discussion can be seen in the supplementary materials.

\subsection{Ablation Study}
\paragraph{Component Analysis}
\Cref{tab:component} shows the ablations about the proposed Gradient-Free Aggregation (GFA) and Proxy Calibration (PC) strategy. We evaluate their effectiveness on both existing benchmarks and our proposed OpenBench.
It can be seen that with GFA fusing CLIP feature and query embeddings, the model's performance is greatly improved on both seen and unseen semantics. 
Besides, PC can facilitate the model's ability to understand various scenarios by cost-free expansion of the training semantic space.
When applying both, the model achieves the best performance.

\input{tables/component_analysis}

\begin{figure}[t]
    \centering
    \includegraphics[width=0.9\linewidth]{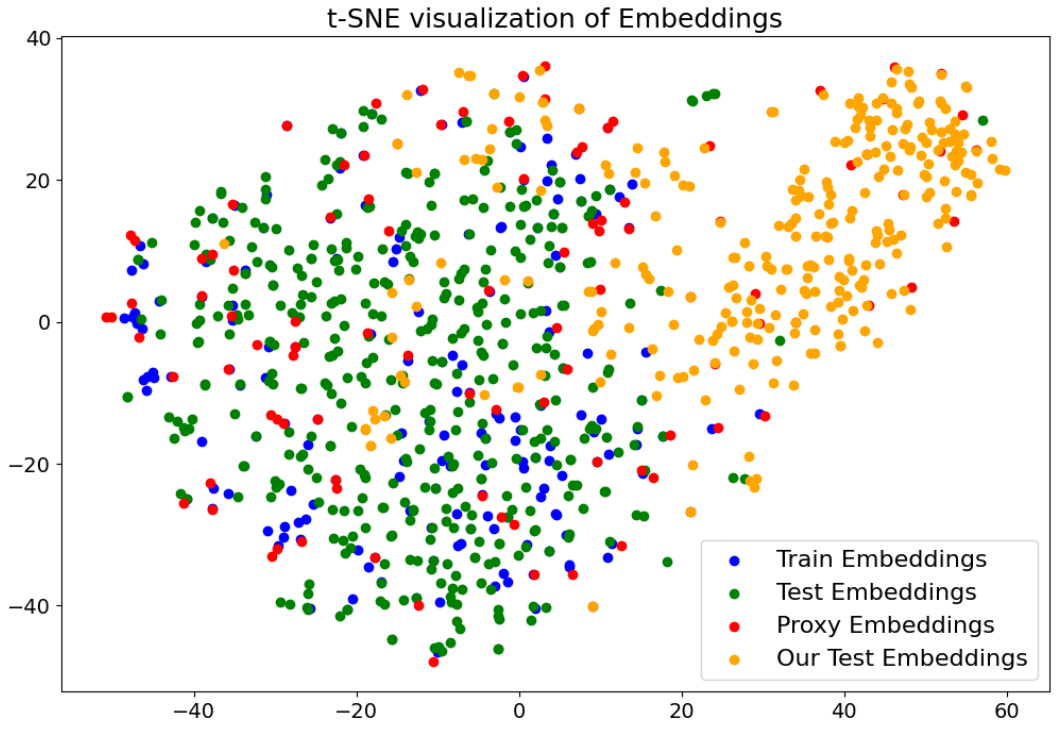}
    \vspace{-5pt}
    \caption{Visualization of category embeddings across the training space, existing evaluation space, our OpenBench space, and the generated proxy space.}
    \label{fig:tsne}
    \vspace{-5pt}
\end{figure}

\vspace{-10pt}
\paragraph{Embedding Distribution Analysis}
To demonstrate that our proposed OpenBench is more open-vocabulary compared to the existing test sets, we perform t-SNE visualization of the category embeddings of the training set, the existing test sets, and our OpenBench in \Cref{fig:tsne}. Besides, we also visualize the proxy embeddings generated by PC.
We can see that the existing test sets (green dots) are overall similar to the training semantics (blue dots). Our OpenBench (orange dots), on the other hand, shows significant differences.
In addition, the proxy embeddings (red dots) generated by PC expand the training set and thus help to improve the robustness and generalization of the model.

\begin{figure}[t]
    \centering
    \includegraphics[width=0.9\linewidth]{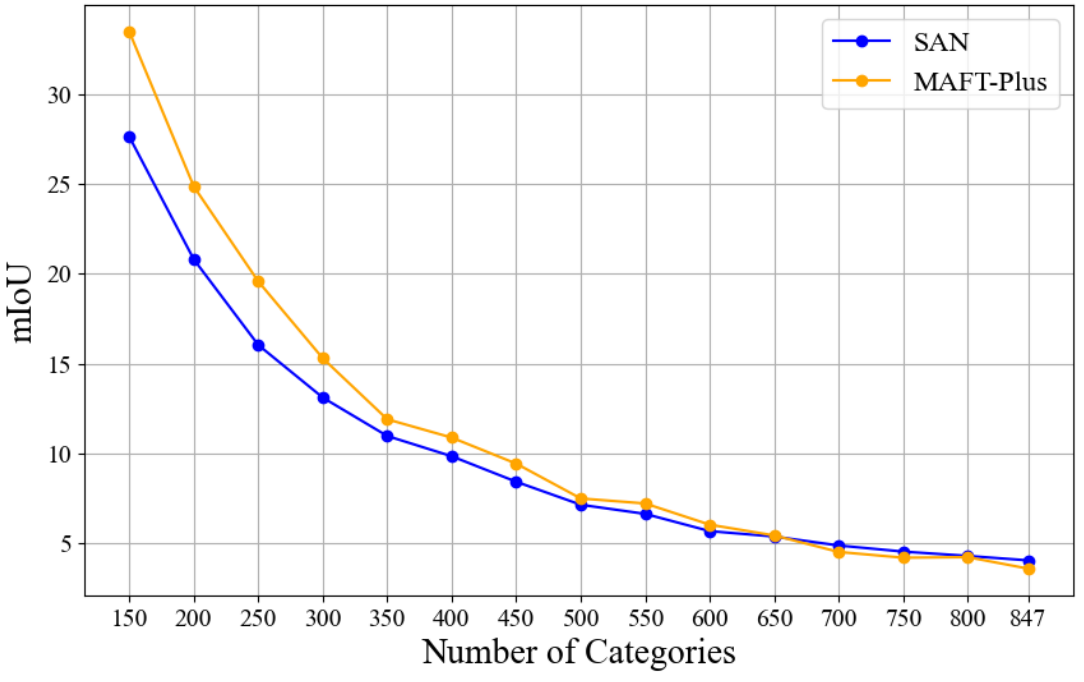}
    \vspace{-10pt}
    \caption{Impact of the number of inference categories on model performance. As the number of irrelevant categories increases, the model's performance on the same image degrades.}
    \label{fig:num_cls}
    \vspace{-5pt}
\end{figure}

\begin{figure*}[t]
    \centering
    \includegraphics[width=\textwidth]{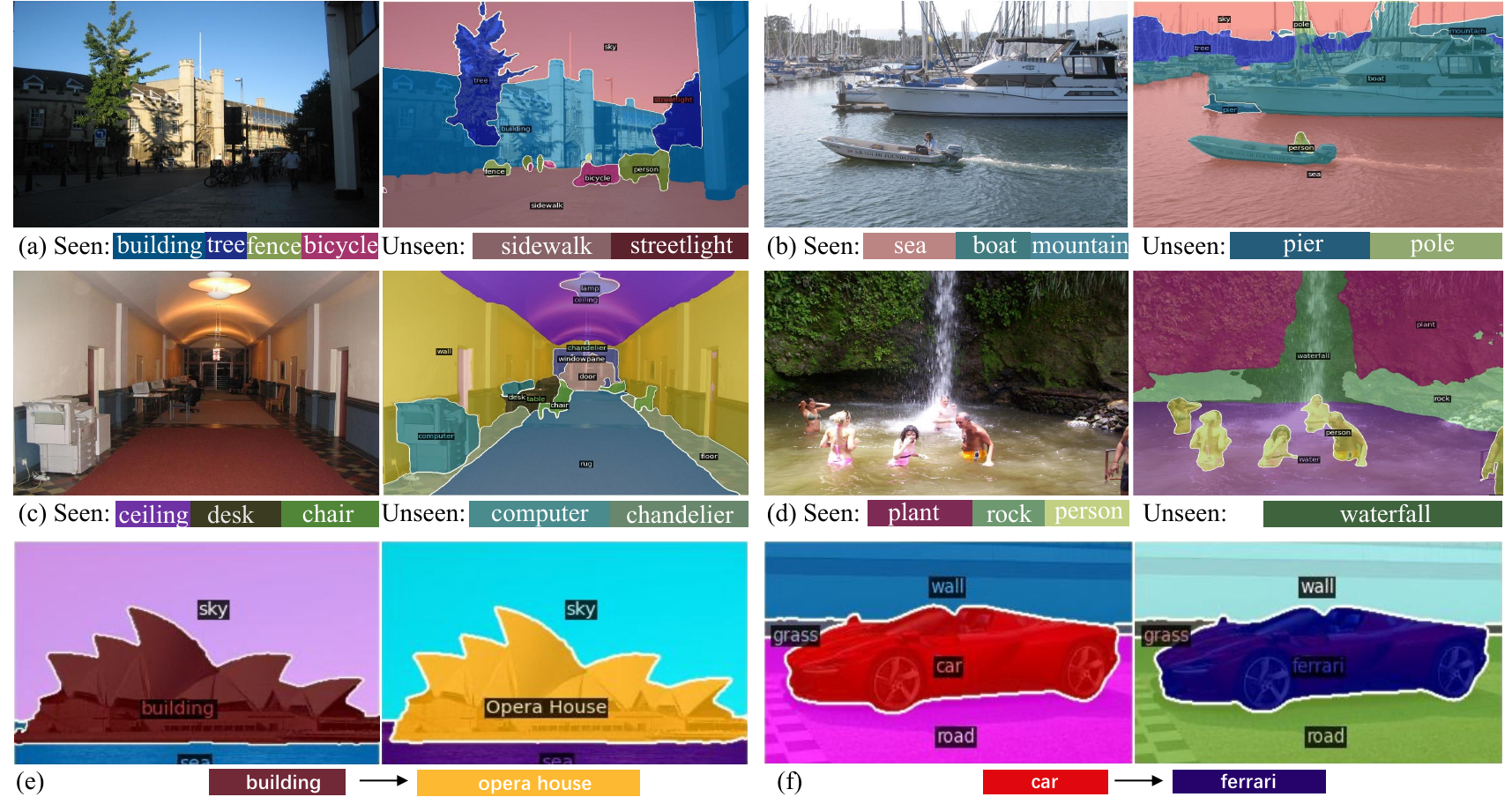}
    \vspace{-18pt}
    \caption{Visualization of segmentation results. (a) - (d) demonstrate the successful segmentation cases of our method for seen and unseen categories. (e) and (f) display the adaptability for flexible text query. Best viewed in color.}
    \label{fig:visual}
    \vspace{-5pt}
\end{figure*}

\paragraph{Analysis about Inference Class Number}
In this part we explore an interesting phenomenon: why the OVS models perform much worse on ADE-847 and PC-459 than on other test sets?
Intuitively, this is because these datasets contain more novel categories. However, we find that another factor significantly impacts model performance, that is the number of categories fed into the model during inference.
Let's start with the conclusion: for the same model and test image, the model performance is worse when there are more inference categories given. In other words, the model may be able to make correct predictions when only a small number of categories are given. However, when more categories are prompted, the model's output will be incorrect.
Here we take two typical methods, SAN~\cite{san} and MAFT-Plus~\cite{maftplus}, as examples and show how their performance on the ADE-150 dataset changes as the number of categories increases.
As shown in \Cref{fig:num_cls}, with the number of given irrelevant categories increases, the models of different technology routes show a consistent decline.

\vspace{-10pt}
\paragraph{Internal Analysis of Proposed Strategies}
In \Cref{tab:inner} we conduct more experiments about the proposed methods.
Specifically, we first compare our gradient-free aggregation designs with the typical learning-based feature fusion approaches in (a). Results show that vanilla attention performs inferior than our GFA, especially on the more open-vocabulary set OpenBench (degraded by more than 2.5 mIoU). 
This is because learning-based methods tend to develop a reliance on the segmentation query embeddings during training, while overlooking the CLIP features that inherently exhibit stronger generalization capabilities, thereby introducing biases.

We also experiment with the influence of different $\gamma$ in Beta distribution to the performance of proxy calibration. The results are shown in \Cref{tab:inner} (b), where $\gamma=1$ is equivalent to do not utilize PC. 
We find that different $\gamma$ benefit the model performance while the performance is better when the intermediate probability density is high ($\gamma=2$).

\input{tables/inner}

\subsection{Visualization}
\Cref{fig:visual} shows some segmentation cases of our OVSNet. It can be seen that our method achieves excellent segmentation performance on various scenarios. Specifically, (a) - (d) demonstrate the excellent segmentation of our method for seen and unseen categories. (e) and (f) display the adaptability for flexible text query, \eg, change ``building" to ``opera house" and ``car" to ``ferrari".

\section{Conclusion}
In this paper, we identify that the existing open-vocabulary segmentation evaluation sets suffer from excessive semantic similarity to the training set, which limits their ability to comprehensively assess the model's open vocabulary understanding capabilities. To address this, we propose a new benchmark, OpenBench, which features significant differences from the training set and includes more fine-grained categories. Comparison on OpenBench further demonstrates that maintaining the generalization space of CLIP is crucial for OVS. Additionally, we introduce OVSNet, a method that leverages proxy calibration for cost-free augmentation of the training space and gradient-free aggregation for fusing heterogeneous features. Experiments show that the proposed OVSNet achieves excellent performance on both existing datasets and the newly proposed OpenBench. We hope that our benchmark and model design could inspire further researches in this area.


%% file: tables/dataset.tex
\begin{table*}[t]
\small
\centering

\caption{Comparison of the statistics of our proposed OpenBench  with existing datasets. `Cls', `Img', and `Sim' indicate class, image, and similarity, respectively. ADE, PC, and VOC denote ADE20K~\cite{ade20k}, Pascal Context~\cite{pascal}, and Pascal VOC~\cite{pascal-voc}}
\vspace{-5pt}
\setlength{\tabcolsep}{3.3pt}
\renewcommand\arraystretch{1.2}
\begin{tabular}{l|c|c|c|c|c|c|c|c}
\hline
\multirow{1}{*}{Dataset}
& Cls Num. &Img Num. &Mean Sim. &Median Sim. &Min Sim. & Max Sim. &Fine Granularity &Semantic Duplication  \\
\hline
VOC~\cite{pascal-voc}    & 20 & 1449 & 0.9756  &1.0000 & 0.8449 & 1.0000 &\texttimes &\texttimes   \\
PC-59~\cite{pascal}    & 59 & 5105 & 0.9481 & 1.0000 &0.7658 & 1.0000 &\texttimes &\texttimes   \\
ADE-150~\cite{ade20k}    & 150 & 2000 & 0.8086 & 0.8228 &0.1960 &1.0000 &\texttimes &\checkmark   \\
Cityscapes~\cite{cityscape}    &33 & 100 & 0.8488 & 0.8370 & 0.6746 &1.0000 &\texttimes &\texttimes   \\
PC-459~\cite{pascal}    & 459 & 5105 &0.8374 &0.8124 &0.6718 &1.0000 &\checkmark &\checkmark   \\
ADE-847~\cite{ade20k}    & 847 & 2000 &0.7926 &0.7913 &0.1960 &1.0000 &\checkmark &\checkmark   \\
\hline
OpenBench (Ours)    & 286 & 6056 & 0.6142 &0.6452 & 0.2608 & 0.7947 &\checkmark &\texttimes   \\
\hline
\end{tabular}
\label{tab:dataset_show}
\vspace{-10pt}
\end{table*}

%% file: tables/main_results.tex
\begin{table*}[t]
\small
\centering
\caption{Performance comparison with state-of-the-art methods on existing benchmarks and the proposed OpenBench. ADE, PC, and VOC denote ADE20K~\cite{ade20k}, Pascal Context~\cite{pascal}, and Pascal VOC~\cite{pascal-voc}, respectively. 171 and 133 indicate the number  of training categories. * denotes the re-trained version according to the official code for fair comparison.}
\vspace{-5pt}
\renewcommand\arraystretch{0.95}
\setlength{\tabcolsep}{7.3pt}
\begin{tabular}{l|l|c|c|c|c|c|c|c}
\toprule
\multirow{1}{*}{Method}  & \multirow{1}{*}{Training Dataset} 
& ADE-150 & ADE-847 & PC-59  & PC-459  & VOC  &OpenBench &Average Score\\
\hline
\multicolumn{9}{c}{ Base-level CLIP Backbone} \\ \hline
SimSeg*~\cite{Simbaseline}  & COCO-Stuff-171  & 21.1 & 6.9  & 51.9 & 9.7     & 91.8  &19.4 &33.5\\
OVSeg~\cite{adapt-mask}    & COCO-Stuff-171   & 24.8 & 7.1    & 53.3 & 11.0 & 92.6 &24.9 &35.6\\
SAN~\cite{san}    & COCO-Stuff-171  & 27.5 &10.1   &53.8 &12.6     &94.0  &39.6 &39.6\\
SCAN~\cite{scan}     & COCO-Stuff-171   &30.8 &10.8   &58.4  &13.2     &\textbf{97.0} &38.7 &41.5\\
CATSeg~\cite{catseg}   & COCO-Stuff-171   &31.8 &12.0 &57.5 &19.0  &94.6 &36.1 &41.8\\
SED~\cite{sed}   & COCO-Stuff-171   & 31.6 & 11.4 & 57.3 & 18.6  &94.4 &33.3 &41.1\\
FCCLIP*~\cite{fcclip}   & COCO-Panoptic-133   & 31.1 & 13.5 & 54.8 & 12.8  &93.2 &40.3 &40.9\\
MAFT~\cite{maft}   & COCO-Panoptic-133   & 29.1 & 10.1 & 53.5  & 12.8  & 90.0 &37.5 &38.8\\
MAFT+~\cite{maftplus}   & COCO-Panoptic-133   &34.6 & 13.8 & 57.5 & 16.2  &95.4 &43.7 &43.5\\
OVSNet (Ours)   & COCO-Panoptic-133    &\textbf{35.8} &\textbf{14.5} &\textbf{58.6} &\textbf{19.1} &95.7 &\textbf{44.9}  &\textbf{44.8}\\
\hline
\multicolumn{9}{c}{ Large-level CLIP Backbone} \\ \hline
SimSeg*~\cite{Simbaseline}  & COCO-Stuff-171  & 21.7 & 7.1  & 52.2 & 10.2     & 92.3  &27.8 &35.2\\
OVSeg~\cite{adapt-mask}  & COCO-Stuff-171   & 29.6 &9.0   &55.7 &12.4     &94.5 &35.6 &39.5\\
SAN~\cite{san}    & COCO-Stuff-171  & 32.1 &12.4   &57.7 &15.7     &94.6  &43.4 &42.7\\
SCAN~\cite{scan} & COCO-Stuff-171   &33.5 &14.0   &59.3  &16.3     &\textbf{97.1} &49.4 &44.9\\
SED~\cite{sed}   & COCO-Stuff-171   & 35.2 & 13.9 & 60.6 & 22.6  &96.1 &45.6 &45.6\\
FCCLIP~\cite{fcclip}   & COCO-Panoptic-133   & 34.0 & 14.8 & 58.4 & 18.2  &95.4 &44.8 &44.3\\
MAFT~\cite{maft}   & COCO-Panoptic-133   & 34.4 & 13.1 & 57.5  & 17.0  & 93.0 &41.7 &42.8\\
MAFT+~\cite{maftplus}   & COCO-Panoptic-133   &36.1 & 15.1 & 59.4 & 21.6  &96.5 &47.3 &46.0\\
OVSNet (Ours)   & COCO-Panoptic-133    &\textbf{37.1} &\textbf{16.2} &\textbf{62.0} &\textbf{23.5 }&96.9 &\textbf{48.6}  &\textbf{47.4}\\
\bottomrule
\end{tabular}
\label{tab:main_result}
\vspace{-10pt}
\end{table*}

%% file: tables/component_analysis.tex
\begin{table}
    \centering
    \small
    \caption{Ablation experiments about the proposed methods. GFA and PC indicate the Gradient-Free Aggregation and Proxy Calibration strategies, respectively.}
    \vspace{-7pt}
    \setlength{\tabcolsep}{9.7pt}
    \begin{tabular}{l|ccc}
    \hline
         Method  &ADE-150  &PC-459  &OpenBench \\
         \hline
         Baseline  &33.1  &14.3  &42.3 \\
         + GFA  &34.7$\uparrow(\textbf{1.6})$  &16.0$\uparrow(\textbf{1.7})$  &43.7$\uparrow(\textbf{1.4})$ \\
         + PC  &33.9$\uparrow(\textbf{0.8})$  &17.2$\uparrow(\textbf{2.9})$  &44.3$\uparrow(\textbf{2.0})$ \\
         + Both  &35.8$\uparrow(\textbf{2.7})$  &19.1$\uparrow(\textbf{4.8})$  &44.9$\uparrow(\textbf{2.6})$ \\
         \hline
    \end{tabular}
    
    \label{tab:component}
    \vspace{-10pt}
\end{table}

%% file: tables/inner.tex
\begin{table}[t]
    \small
    \centering
    \caption{Ablation experiments within each modules. $(a)$ shows the comparison of our gradient-free algorithm with learning-based method, \textit{e.g.}, cross attention. $(b)$ demonstrates the effect of $\gamma$ of Beta distribution in the proxy learning strategy.}
    \vspace{-5pt}
    \renewcommand\arraystretch{0.9}
    \renewcommand\tabcolsep{8pt}
    \begin{tabular}{l|c|c|c}
    \hline
        &ADE-150 &PC-459 &OpenBench\\ \hline     
    \multicolumn{4}{l}{ \textit{(a) Comparisons of GFA and learning-based strategy}} \\ \hline
    Self Attn   &34.0 &14.7     &40.6    \\  
    Cross Attn   &34.2 &14.8     &41.2    \\    
    GFA   &\textbf{34.7}  &\textbf{16.0}  &\textbf{43.7}   \\
    \hline
    \multicolumn{4}{l}{ \textit{(b) Effect of sampling distribution of PC}} \\ \hline
    $\gamma=1$    &33.1 &14.3     &42.3   \\   
    $\gamma=0.5$    &33.6$\uparrow(\textbf{0.5})$ &16.1$\uparrow(\textbf{1.8})$     &43.7$\uparrow(\textbf{1.4})$   \\   
    $\gamma=2$   &\textbf{33.9}$\uparrow(\textbf{0.8})$  &\textbf{17.2}$\uparrow(\textbf{2.9})$  &\textbf{44.3}$\uparrow(\textbf{2.0})$ \\
    \hline
    \end{tabular}
    \label{tab:inner}
    \vspace{-10pt}
\end{table}